\pdfoutput=1
\documentclass[11pt]{article}
\usepackage[utf8]{inputenc}
\usepackage{newunicodechar}
\newunicodechar{Ḥ}{\.{H}}
\usepackage[final]{acl}

\usepackage{times}
\usepackage{latexsym}

\usepackage[T1]{fontenc}
\usepackage[utf8]{inputenc}
\usepackage{authblk}

\usepackage{microtype}

\usepackage{inconsolata}
\usepackage{url}
\usepackage{graphicx}
\usepackage{booktabs}
\usepackage{pifont}
\usepackage{newunicodechar}
\newunicodechar{✓}{\ding{51}}
\newunicodechar{✗}{\ding{55}}
\usepackage[inline]{enumitem}

\title{Transformer Tafsir at QIAS 2025 Shared Task: Hybrid Retrieval-Augmented Generation for Islamic Knowledge Question Answering}

\author{
  \textbf{Muhammad Abu Ahmad}\textsuperscript{1},
  \textbf{Mohamad Ballout}\textsuperscript{1},
  \textbf{Raia Abu Ahmad}\textsuperscript{2},
  \textbf{Elia Bruni}\textsuperscript{1}
}
\affil{
    \textsuperscript{1}Institute of Cognitive Science, University of Osnabrück, Osnabrück, Germany \\
    \textsuperscript{2}Deutsches Forschungszentrum für Künstliche Intelligenz GmbH (DFKI), Berlin, Germany \\
    \small Corresponding author: \href{mailto:mabuahmad@uni-osnabrueck.de}{mabuahmad@uni-osnabrueck.de}
}

\begin{document}
\maketitle
\begin{abstract}
This paper presents our submission to the QIAS 2025 shared task on Islamic knowledge understanding and reasoning. We developed a hybrid retrieval-augmented generation (RAG) system that combines sparse and dense retrieval methods with cross-encoder reranking to improve large language model (LLM) performance. Our three-stage pipeline incorporates BM25 for initial retrieval, a dense embedding retrieval model for semantic matching, and cross-encoder reranking for precise content retrieval. We evaluate our approach on both subtasks using two LLMs, Fanar and Mistral, demonstrating that the proposed RAG pipeline enhances performance across both, with accuracy improvements up to 25\%, depending on the task and model configuration. Our best configuration is achieved with Fanar, yielding accuracy scores of 45\% in Subtask 1 and 80\% in Subtask 2.
\end{abstract}

\section{Introduction}

QIAS 2025 is a question answering (QA) shared task that aims to evaluate large language models' (LLMs) ability to understand and reason within Islamic knowledge~\cite{qias2025,bouchekif2025islamic}. The task is divided into two subtasks: (1) Islamic Inheritance Reasoning, requiring precise application of inheritance law principles, and (2) Islamic Assessment, covering general Islamic knowledge across different topics such as theology, jurisprudence, biography, and ethics. Islamic jurisprudence (Fiqh) and inheritance law ('Ilm al-Mawārīth) present unique challenges within natural language processing (NLP) in the Arabic language, as it highlights the differences between Modern Standard Arabic (MSA) and Classical Arabic used in religion-related texts.

\begin{figure*}[t]
\center
  \includegraphics[width=0.9\linewidth]{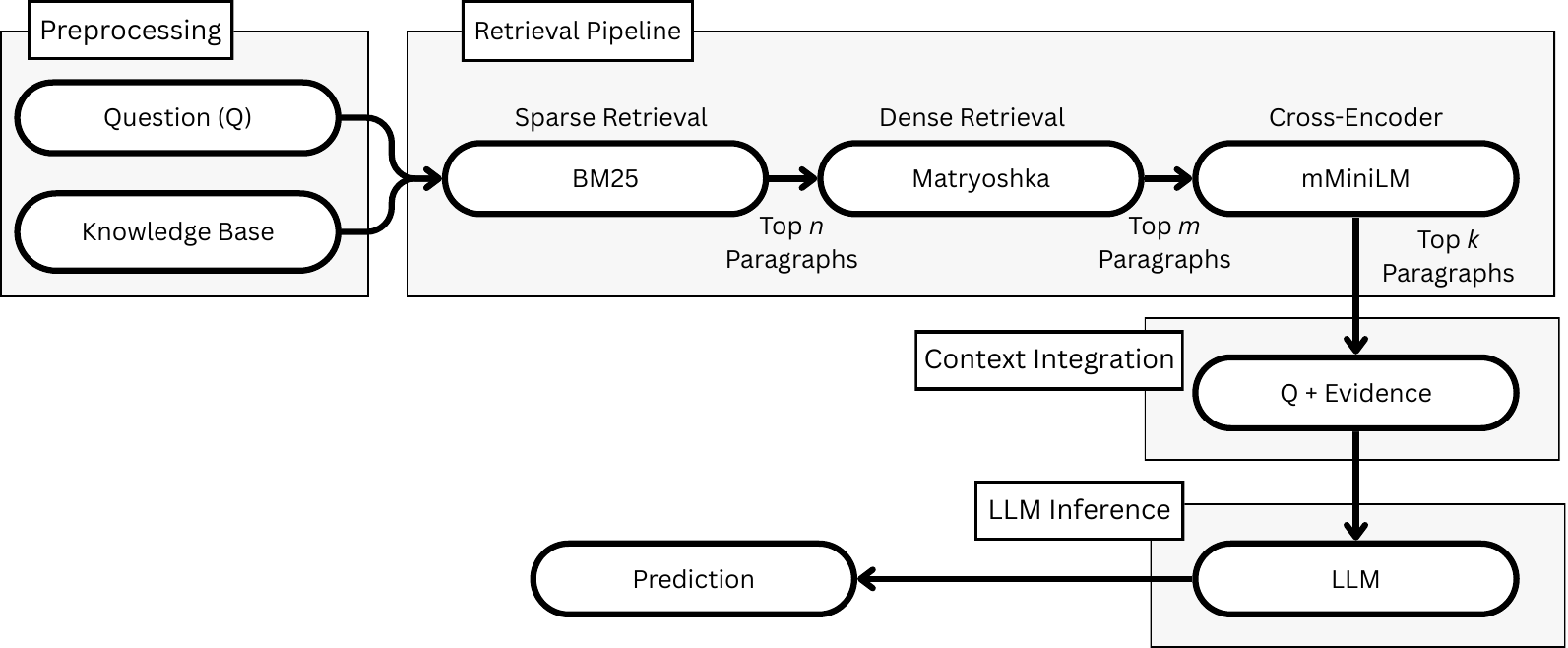}
  \caption {Proposed retrieval-augmented pipeline for the QIAS shared task.}
  \label{fig:approach}
\end{figure*}

We tackle the QIAS shared task using a hybrid, naive retrieval-augmented generation (RAG) pipeline specifically designed for Arabic Islamic knowledge. As shown in Figure~\ref{fig:approach}, our approach consists of four components: \begin{enumerate*}
    \item Preprocessing; \item Three-stage hybrid retrieval pipeline; \item Context integration; and \item LLM inference. 
\end{enumerate*} Our suggested retrieval system combines sparse retrieval via BM25 \cite{robertson2009probabilistic}, dense retrieval using Arabic-optimized embeddings \cite{nacar2025gate}, and a miniLM-based cross-encoder reranking model for final passage selection. 

In this paper, we present our system in detail and discuss our main contributions, including a specialized Arabic preprocessing pipeline, a hybrid three-stage retrieval architecture that combines complementary retrieval methods, a comprehensive evaluation across multiple LLMs demonstrating consistent improvements RAG context integration, and an analysis of the pipeline's performance. To facilitate reproducibility, we make our implementation publicly available.\footnote{\url{https://gitlab.com/mhauesh/qias-shared-task-2025-solution-implementation}}

\section{Background}

\textbf{Task Setup.} The QIAS 2025 shared task features two scholar-verified multiple-choice question (MCQ) subtasks with three difficulty levels. Subtask 1 (Islamic Inheritance Reasoning) covers 'Ilm al-Mawārīth with 9,450 training, 1,500 validation, and 1,000 test questions, plus 32,000 fatwas. Subtask 2 (Islamic General Knowledge) tests general Islamic knowledge from a selection of 25 classical Islamic knowledge books with 800 validation and 1,000 test questions. The complete corpus of books was also provided by the organizers. 

\textbf{Related Work.} QA tasks have demonstrated significant benefits from retrieval-augmented pipelines, particularly when domain-specific knowledge bases exist~\cite{arslan2024survey}. RAG combines retrieval with generative models by first retrieving relevant passages from a knowledge base, then feeding them as context to language models for answer generation~\cite{gao2023retrieval}.

Modern hybrid retrieval pipelines typically consist of three stages: sparse retrieval, dense retrieval, and cross-encoder reranking~\cite{huyen2024ai}. BM25, a lexical retrieval method using TF-IDF and document-length normalization, serves as the most widely-used sparse retriever. Dense retrievers embed queries and passages into shared vector spaces using transformer models~\cite{karpukhin2020dense}, capturing semantic relationships that lexical methods might miss. Finally, cross-encoders provide higher precision by jointly scoring query-document pairs with full attention~\cite{cheng-etal-2023-task}.

Recent Arabic-focused RAG research demonstrates the value of this approach. For example, Arabica QA~\cite{abdallah2024arabicaqa} presents QA pairs with a dense retrieval model pre-trained on Arabic for open-domain QA, proving the effectiveness of RAG pipelines on performances of various LLMs. Similarly, \citet{al-rasheed-etal-2025-evaluating} show that integrating RAG pipelines improves LLM prediction results over retrieval-free setups.

Recent advancements in Arabic embedding models~\cite{nacar2025gate}, have significantly improved representation quality for retrieval tasks, with custom Matryoshka embeddings performing highly on the MTEB leaderboard.\footnote{\url{https://huggingface.co/spaces/mteb/leaderboard}} These models offer compact yet powerful embeddings well-suited for scalable retrieval and reranking in Arabic-language RAG pipelines. In the context of Islamic and Quranic QA, however, additional challenges arise due to the linguistic divergence between MSA queries and Classical Arabic source texts, semantic ambiguity in Quranic language, and the scarcity of high-quality, domain-specific datasets~\cite{oshallah2025cross}. Prior work has demonstrated the potential of retrieval-augmented QA systems~\cite{khalila2025investigating}, showing that even small LLMs can generate relevant and faithful answers when grounded in appropriate retrieval context. Other studies have explored individual components, such as dense retrievers and rerankers for Arabic QA~\cite{alsubhi2025optimizing, el2024exploring}. 

Building on this foundation, our work contributes a task-specific, modular retrieval architecture that combines sparse, dense, and reranking components in a unified pipeline tailored for Classical Arabic QA in the domain of Islamic knowledge, providing insight into the utility of a hybrid retrieval system under realistic conditions.

\section{System Overview}

The architecture of the pipeline consists of four main components: Arabic text preprocessing and knowledge base construction, three-stage hybrid retrieval, context integration, and LLM inference.

\textbf{Arabic Text Preprocessing.} Arabic text processing poses unique challenges due to rich morphology, orthographic variations, and diacritical marks~\cite{habash2010introduction}. Since each component in our retrieval pipeline requires specific preprocessing needs, we implement a two-tier preprocessing approach: \emph{Full preprocessing} and \emph{light preprocessing}. The former is used for BM25 indexing, and includes stopword removal using enhanced NLTK \cite{bird-loper-2004-nltk} lists, tokenization via CAMeL tools \cite{obeid-etal-2020-camel}, and token filtering by length and content criteria. On the other hand, light preprocessing is used for dense retrieval and cross-encoder inputs and includes formatting normalization, punctuation removal, citation removal, character normalization, and dediacritization using CAMeL tools. The important distinguishing factor between the two preprocessing approaches is preserving semantic information when it comes to dense retrieval pipelines. 

\begin{table}[ht] 
\centering
\small
\begin{tabular}[width=\columnwidth]{lcc}
\toprule
 & Light& Full\\
 \midrule 
 Formatting & \ding{51} & \ding{51}   \\
 Punctuation removal & \ding{51} & \ding{51} \\
 Dediacritization & \ding{51} & \ding{51} \\
 Character normalization  & \ding{51} & \ding{51} \\
 Citations removal & \ding{51} & \ding{51} \\
 Stopwords removal & \ding{55} & \ding{51} \\
 Tokenization & \ding{55} & \ding{51} \\
\bottomrule
\end{tabular}
\caption{Preprocessing procedure for retrieval methods.}
\label{tab:preprocessing}
\end{table} 

\textbf{Knowledge Base Construction.} We construct domain-specific knowledge bases from the provided training materials: For Subtask 1, we process 32,000 IslamWeb fatwas in JSON format. Each fatwa contains structured fields including category, question, answer, and metadata. We treat each complete fatwa as a single retrieval unit to maintain contextual coherence. For Subtask 2, we process classical Islamic books provided in HTML and DOCX formats, implementing paragraph-based chunking and applying overlap strategies to prevent information loss. We create tri-directional mappings between fully processed chunks, lightly processed chunks, and original text, enabling seamless integration across different retrieval stages.

\textbf{Retrieval Pipeline.} Our hybrid retrieval system combines complementary retrieval methods: \begin{enumerate*}
    \item Sparse retrieval using BM25, which provides initial candidate selection using lexical matching using the fully preprocessed Arabic text, retrieving the top 1000 candidates.
    \item Dense retrieval, applying semantic matching using Arabic-optimized embedding models. Based on existing benchmarks \cite{enevoldsen2025mmtebmassivemultilingualtext}, we embed the lightly preprocessed text using Arabic-Triplet-Matryoshka-V2, selecting the top 200 passages closest to the question based on cosine similarity.
    \item Cross-encoder reranking, which provides final precision enhancement using transformer models trained for relevance scoring. The cross-encoder jointly processes query-passage pairs, allowing full attention across inputs for more accurate relevance assessment. We use a miniLMv2 model fine-tuned on the MMARCO dataset.\footnote{\url{https://huggingface.co/cross-encoder/mmarco-mMiniLMv2-L12-H384-v1}} Due to context window limitations in the used LLMs, we retrieve the top 5 passages with associated relevance scores for context integration in the LLM prompt.
\end{enumerate*}

\textbf{Context Integration.} We designed a prompt to support both tasks based on prior research \cite{schulhoff2024prompt}. First, we defined a domain-specific persona who is an expert Islamic scholar. Then, for each question, we retrieved relevant context passages and integrated them into the prompt using a format that prioritizes them as sources. We also added few-shot examples by selecting two random questions from the development set, demonstrating correct reasoning patterns, and included format constraints to enforce valid multiple-choice responses. We include the full prompt template in Appendix~\ref{sec:appendix}.

\section{Experimental Setup}

We evaluate our proposed system across two distinct LLMs representing different model families and access patterns. First, we use Fanar \cite{team2025fanar} via its API, which is a specialized Arabic LLM designed for Islamic content. Then, we experiment with Mistral (specifically, mistral-saba-24b),\footnote{\url{https://mistral.ai/news/mistral-saba}} a state-of-the-art open-weight model, accessed through the Groq API.\footnote{\url{https://console.groq.com/home}} These LLMs were chosen since they were made available by the organizers of the QIAS shared task.

We use the same retrieval configuration on all tested LLMs: From BM25, we retrieve the top 1000 most relevant passages from the knowledge base, from dense retrieval, we filter those to the top 200 passages most similar to the given question, and finally, from the cross-encoder, we retrieve the top 5 passages to be integrated as context when prompting LLMs. We process the query and the knowledge base using CAMeL Tools (v1.2.0) for normalization, dediacritization, and tokenization. Additionally, we implement custom routines for citation removal, formatting cleanup, and chunking the knowledge base.

Performance of LLMs is measured using accuracy, which is the percentage of questions where the model's prediction exactly matches the correct answer, evaluated during the testing phase of the shared task via the provided platform.

\section{Results \& Discussion}

We present our results in Table~\ref{tab:results} for both subtasks, demonstrating varying levels of improvement when incorporating RAG across all tested configurations. The magnitude of improvement varies significantly between tasks and models. Subtask 1 (Islamic Inheritance) shows modest but consistent improvements of 1\%-4\% when implementing our proposed pipeline. We hypothesize that the specialized nature of inheritance law calculations may limit RAG effectiveness, as these problems often require precise mathematical reasoning rather than factual retrieval, as well as the high degree of semantic similarity between questions that require different methods to solve. However, we note that Subtask 2 (General Islamic Knowledge) exhibits substantial performance boosts with the RAG pipeline, with improvements ranging from 10\%-25\%. This dramatic enhancement suggests that general Islamic knowledge questions benefit significantly from access to authoritative source material.

\begin{table}[ht!]
\centering
\small
\begin{tabular}{lc}
\toprule
\textbf{Model Configuration} & \textbf{Accuracy Score}\\
 \midrule
 \multicolumn{2}{c}{\emph{Subtask 1: Islamic Inheritance}} \\ \midrule
Fanar Baseline  & 44.0\%\\
\textbf{Fanar Transformer Tafsir}  & \textbf{45.0\%}\\
\midrule
Mistral Baseline  & 35.0\%\\
Mistral Transformer Tafsir  & 39.0\%\\
\midrule
 \multicolumn{2}{c}{\emph{Subtask 2: General Islamic Knowledge}} \\ \midrule
Fanar Baseline  & 55.0\%\\
\textbf{Fanar Transformer Tafsir}  & \textbf{80.0\%}\\
\midrule
Mistral Baseline  & 69.0\%\\
Mistral Transformer Tafsir  & 79.0\%\\
\bottomrule
\end{tabular}
\caption{Results on the given test sets for both subtasks of the QIAS shared task using our proposed hybrid retrieval pipeline (+RAG) compared to baseline model performance (without RAG).}
\label{tab:results}
\end{table}

Comparing different LLMs, we note that Fanar shows smaller relative improvements (1\%-25\%) but reaches higher absolute performance on Subtask 2, likely due to its Islamic domain specialization and prior exposure to similar training data. On the other hand, Mistral demonstrates more consistent relative improvements (4\%-10\%) across both tasks, suggesting the overall benefit of RAG pipelines to improve performance of models on very specific domains, such as Islamic knowledge in this case, when similar in-domain data was most likely lacking in their training processes. 

\section{Error Analysis}

A manual error analysis reveals three key patterns. First, models struggle with the complex, fractional reasoning in Subtask 1, indicating a need for symbolic reasoning beyond current RAG approaches. Second, errors arise when retrieved context is relevant but incomplete, highlighting the importance of comprehensive knowledge bases and accurate retrieval. Finally, some questions demand logical inference that simple retrieval cannot solve, suggesting a need for specialized training methodologies \cite{ke2025survey} or reasoning-based prompting \cite{qiao-etal-2023-reasoning}.

Analyzing by difficulty level for our best results (Fanar Transformer Tafsir for both tasks), we see varied performance by the LLMs on the two subtasks. Task 1 declined from 54.27\% on beginner questions to 36.22\% on advanced questions, while Subtask 2 showed a similar pattern (82.21\% to 73.33\%), as shown in Table~\ref{tab:performance_matrix}. Subtask 2 errors were more evenly distributed, but showed difficulty in theological reasoning and jurisprudential methodology.

\begin{table}[ht]
\centering
\resizebox{\columnwidth}{!}{
\begin{tabular}{lcccc}
\toprule
\textbf{Task} & \textbf{Beginner} & \textbf{Intermediate} & \textbf{Advanced} & \textbf{Overall} \\
\midrule
Subtask 1 & 53.40\% & -- & 36.00\% & 44.70\% \\
Subtask 2 & 81.86\% & 78.67\% & 73.33\% & 80.10\% \\
\bottomrule
\end{tabular}
}
\caption{System performance by task and difficulty level.}
\label{tab:performance_matrix}
\end{table}

\section{Conclusion}

We presented a hybrid RAG pipeline for the QIAS 2025 shared task on Islamic knowledge QA. Our pipeline involved Arabic-specific preprocessing, a three-stage retrieval architecture (BM25, dense retrieval, cross-encoder reranking), context integration, and LLM inference. Based on evaluations of Fanar and Mistral, we showed that our method consistently outperformed baselines, demonstrating that RAG improves accuracy, especially for general knowledge over structured inheritance problems. To improve the presented system, future work can explore dynamic context selection, domain fine-tuning, and integrating structured reasoning modules for inheritance law.

\section*{Limitations} 

Our system struggles with complex inheritance problems requiring multi-step mathematical reasoning, as it lacks symbolic reasoning capabilities. The presented system is not fine-tuned, hence the semantic similarity of recurring words and phrases in subtask 1 limits the system's ability to retrieve precise relevant passages. The system is also dependent on the quality of the external knowledge base and does not explore knowledge curation. We did not use the provided training data, so domain fine-tuning remains unexplored. The number of retrieved passages in each step of the pipeline requires further investigation in order to fully maximize the system's capabilities. Choosing said parameters (n,m and k) was based on trials conducted on the developments sets and not on the test sets. Finally, we did not explore performance on additional LLMs, which maybe have yielded better results, due to time and compute limitations. 

\section*{Acknowledgments}

We thank the QIAS 2025 organizers for providing comprehensive datasets and evaluation frameworks. We also acknowledge computational resources provided by the University of Osnabrück.

\bibliography{acl_latex}

\appendix
\section{Implementation - Technical Details}
\label{sec:appendix}

\subsection{Hyperparameters}

\noindent\textbf{BM25 Configuration:}\\
k1 = 1.2 (term frequency saturation); b = 0.75 (length normalization); Top candidates = 1000

\noindent\textbf{Dense Retrieval:}\\
Embedding dimension = 768; Similarity metric = Cosine similarity; Top candidates = 200; Batch size = 8 for embedding computation

\noindent\textbf{Cross-Encoder Reranking:}\\
Model: Arabic BERT-based reranker; Final candidates = 5; Temperature = 0.1 for stable rankings

\subsection{Prompt}
You are an expert Islamic scholar. Your task is to answer multiple-choice questions.
[Examples...]
First, use the following reference text to determine the answer: 
RAG CONTEXT
QUESTION:
MULTIPLE CHOICES:
Your response MUST be only the single capital letter of the correct option. Do not include 'Answer:', explanations, or any other text.

\noindent\textbf{Chunking Strategy:}\\
Target chunk size: 200 tokens (BM25 optimized); Overlap: 20 tokens between adjacent chunks; Minimum chunk size: 50 tokens; Maximum chunk size: 400 tokens

\end{document}